\documentclass{article}

\usepackage[preprint]{neurips_2023}

\usepackage[numbers]{natbib}
\usepackage[utf8]{inputenc} % allow utf-8 input
\usepackage[T1]{fontenc}    % use 8-bit T1 fonts
\usepackage{hyperref}       % hyperlinks
\usepackage{url}            % simple URL typesetting
\usepackage{booktabs}       % professional-quality tables
\usepackage{amsfonts}       % blackboard math symbols
\usepackage{nicefrac}       % compact symbols for 1/2, etc.
\usepackage{microtype}      % microtypography
\usepackage{xcolor}         % colors
\usepackage{graphicx}
\usepackage{amsmath}
\usepackage{amssymb}
\usepackage[normalem]{ulem}
\usepackage{wrapfig}
\usepackage{subcaption}

\usepackage[capitalize]{cleveref}
\crefname{section}{Sec.}{Secs.}
\Crefname{section}{Section}{Sections}
\Crefname{table}{Table}{Tables}
\crefname{table}{Tab.}{Tabs.}

\title{TransDreamerV3: Implanting Transformer In DreamerV3}

\author{Shruti Sadanand Dongare\\
{\tt\small dshruti20@vt.edu}
\And
Amun Kharel\\
{\tt\small akharel@vt.edu}
\And
Jonathan Samuel\\
{\tt\small samueljon17@vt.edu}
\And
Xiaona Zhou\\
{\tt\small xzhou1@vt.edu}
}

\begin{document}

\maketitle

% \begin{abstract}
% \end{abstract}

\begin{abstract}
This paper introduces TransDreamerV3, a reinforcement learning model that enhances the DreamerV3 architecture by integrating a transformer encoder. The model is designed to improve memory and decision-making capabilities in complex environments. We conducted experiments on Atari-Boxing, Atari-Freeway, Atari-Pong, and Crafter tasks, where TransDreamerV3 demonstrated improved performance over DreamerV3, particularly in the Atari-Freeway and Crafter tasks. While issues in the Minecraft task and limited training across all tasks were noted, TransDreamerV3 displays advancement in world model-based reinforcement learning, leveraging transformer architectures.
\end{abstract}
\section{Introduction}\label{introduction}

% \textcolor{red}{
%     \textit{
%     \begin{itemize}
%         \item Clearly state the goal of project
%         \item Specify the input and desired output
%     \end{itemize}
% }}

Reinforcement learning (RL) has emerged as a powerful paradigm, allowing agents to autonomously learn optimal strategies by interacting with their environments \cite{Russell2020IntroAI}. A particularly noteworthy concept in the RL domain is the idea of World Models \cite{Ha2018WorldM}—a structured representation of the environment that enables agents to simulate potential future outcomes, facilitating learning within an imagined environment. These models have played a crucial role in addressing tasks that require foresight and planning \cite{Ha2018WorldM,Wu2022DayDreamerWM}, thereby bridging the gap between traditional model-free methods, which learn solely from past interactions, and model-based approaches that leverage predictive models for enhanced efficiency and performance \cite{Russell2020IntroAI}.

One of the most notable advancements in the field of reinforcement learning with World Models is the introduction of DreamerV3 \cite{hafner2023mastering}. This state-of-the-art algorithm stands out for its capacity to generalize and master a diverse array of domains using fixed hyperparameters. DreamerV3's approach represents the first algorithm to successfully address the complex challenge of collecting diamonds in the popular video game Minecraft from scratch. While DreamerV3 has achieved significant milestones, it heavily relies on Recurrent State Space Models (RSSM), which combine the strengths of recurrent neural networks (RNN) with structured latent variable models \cite{Hafner2018LearningLD}. Despite the power of RNNs in capturing sequential dependencies, they have historically faced challenges related to memory decay, especially concerning long-term memory retention \cite{Ha2018WorldM}. This inherent limitation creates difficulties in environments where agents need to remember and act upon events from distant past interactions. As tasks become more complex and demand enhanced memory capabilities, there is a clear need to explore alternative approaches.

Transformers, originally introduced for natural language processing, have consistently exhibited significant advantages over other architectures, particularly in memory management and parallel processing \cite{Vaswani2017AttentionIA}. Their incorporation into reinforcement learning has uncovered remarkable potential, especially in tasks requiring substantial memory capabilities  \cite{chen2021decision}. In this context, the TransDreamer  \cite{chen2022transdreamer} architecture comes to the forefront, leveraging the strengths of world models while capitalizing on the transformer's benefits. This architecture builds upon the foundation of the original Dreamer architecture \cite{Hafner2019DreamTC}. At the core of this innovation is the Transformer State-Space Model (TSSM)—a stochastic transformer-based state-space model.

In this research, we introduce TransDreamerV3, an evolution of the DreamerV3 framework. Our model aims to harness the advantages of transformers by integrating components from the TSSM framework of TransDreamer into the DreamerV3 architecture. Through this integration, we hypothesize improvements not only in memory capabilities but also in the overall performance and adaptability of the model across a variety of tasks.

We implemented several modifications in our model, transitioning from RSSM to TSSM. Firstly, for the deterministic state model, we substituted the existing gated recurrent unit with a transformer encoder. This encoder is independent of prior deterministic states but dependent on all representation states and actions. Regarding the representation model, we modified the belief to exclude the deterministic state. For trajectory roll-out, we limited imagined trajectories to three per training sample, and the replay buffer prioritized trajectories with higher rewards. Lastly, we trained the policy by keeping the transformer parameters frozen during training.

While implementing DreamerV3, the team substituted the GRU in the RSSM module with a transformer, utilizing the relatively recent programming languages JAX and Ninjax. Ninjax, constructed on the JAX foundation, facilitated the development of intricate neural network architectures. To ensure harmony with DreamerV3's existing codebase, the team adopted a naive transformer, encountering obstacles due to the scarcity of documentation and community knowledge about Ninjax. The naive transformer features a manually constructed attention mechanism, omits positional encoding and dropout layers, and applies layer normalization after the attention and feed-forward network steps.

We summarize the key contributions of this paper as follows:
\begin{itemize}
    \item  We introduce TransDreamerV3, a world model agent based on transformers, extending the success of TransDreamer and DreamerV3.
    \item We enhanced the DreamerV3 codebase by substituting the RSSM architecture with the Naive TSSM architecture, developed using the modern programming languages JAX and Ninjax. 
\end{itemize}

\section{Related Work} \label{related_work}
\subsection{World Models}
In the context of World Models, DreamerV3 is a notable advancement, improving upon DreamerV2 \cite{Hafner2020MasteringAW} in several ways. It introduces discrete regression for the critic, enhancing learning efficiency in environments with sparse rewards \cite{Hafner2020MasteringAW}. The actor network in DreamerV3 efficiently maximizes returns \& maintains exploration balance within different reward conditions. This model has shown success in over 150 tasks demonstrating its wide applicability, generalizability and robust performance with fixed hyperparameters \cite{Hafner2020MasteringAW}. DreamerV3 achieved the groundbreaking feat of autonomously collecting diamonds in the MineCraft environment, without any human guidance. Despite DreamerV3's success, its performance is not entirely consistent across different tasks, as it only occasionally solves the Minecraft diamond challenge, indicating limitations in long horizon tasks possibly due to a lack of long-term memory retention in the Recurrent State Space Model (RSSM) framework. 

\subsection{Transformers}
Transformers find applications in a wide array of tasks, yet within the realm of Reinforcement Learning (RL), they exhibit unique emerging potential, alongside some distinct challenges from an embodied AI perspective. Prior works \cite{parisotto2020stabilizing}, \cite{irie2021going} have tackled the integration of transformers in RL, often employing strategies like introducing gating layers on top of transformer layers to enhance training stability. Diverse approaches are explored across different works; for instance, the 'Decision Transformer' \cite{chen2021decision}, and work like \cite{janner2021offline} represent a concurrent sequence-modeling approach, wherein it re-frames the RL problem as a sequence modeling task and employs transformers to predict actions, eliminating the need for additional networks for actor or critic roles. 

\subsection{World Models With Transformers}
IRIS \cite{Micheli2022TransformersAS} employs a discrete autoencoder \& an autoregressive Transformer as the World model framework, demonstrating great sample efficiency in complex environments like the Atari 100k benchmark, outperforming humans in 10 out of 26 games with limited data. Despite its efficiency, it encounters challenges in tasks requiring extensive memory. TransDreamer \cite{chen2022transdreamer} introduces the Transformer State-Space Model (TSSM), specifically designed to handle long-range memory access and memory-based reasoning in both 2D and 3D visual RL tasks. It addresses the limitations of RSSM frameworks employed within DreamerV2's architecture and demonstrates improved performance in complex tasks requiring sophisticated temporal understanding.

\section{Methods}\label{methods}

Transformers have been shown to excel in tasks requiring complex long-term temporal dependencies such as memory based reasoning and learning complex interactions between historical states  \cite{chen2022transdreamer},\cite{ritter2021rapid},\cite{banino2020memo}. These are important and desirable capabilities that a World Model needs. We hypothesize that a Transformer based World Model in DreamerV3 will outperform RNN-based Dreamer agent for tasks requiring complex and long-term memory dependency. 

We use two main papers \cite{hafner2023mastering} and  \cite{chen2022transdreamer} to replace the backbone of the world model used in Dreamer, which is a stochastic recurrent neural network called Recurrent State-Space Model (RSSM) by a Transformer State-Space Model (TSSM).

\subsection{Transformer State Space Model (TSSM) }
We get the idea of TSSM as shown from the paper  \cite{chen2022transdreamer}. We believe using TSSM instead of RSSM in DreamerV3 would increase efficiency for various tasks such as Atari 100K, Proprio Control, BSuite, Visual Control, Atari 200M and Crafter. Because of its ability to do complex interactions and learning long-term dependencies, it would also perform the Minecraft Diamond task faster and more efficiently. TSSM has following desirable qualities based on paper  \cite{chen2022transdreamer}: i) Direct access to past states, ii) Can update each time step in parallel during training, iii) Ability to roll out sequentially for trajectory imagination for test time, and iv) Stochasticity for latent variable. We are going to make two primary contributions as stated in the Problem Statement using the TSSM backbone. 

\begin{figure}[h]
\centering
\includegraphics[scale=0.4]{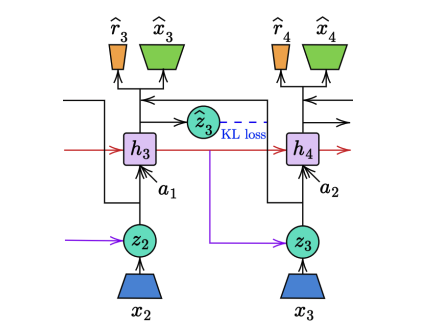}
\caption{RSSM model where red arrow makes sequential computational necessary}
\end{figure}

\begin{figure}[h]
\centering
\includegraphics[scale=0.4]{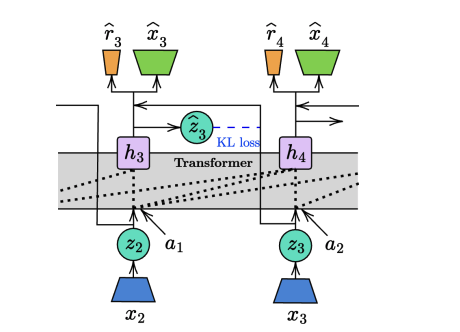}
\caption{Modified TSSM model by eliminating deterministic state dependency (red arrows depicted in Figure 1), facilitating parallel updates across all time steps}
\end{figure}

\subsection{Architectural changes from RSSM to TSSM }
Figure 1 shows the architecture of RSSM and Figure 2 depicts the architecture of TSSM, which will be used to update the backbone of Dreamer-v3. RSSM has sequentially dependent computation when we have to update state \(h_t = f_{gru} (h_{t-1}, z_{t-1}, q_{t-1},) \), which is shown in red arrows in Figure 1. TSSM removes this sequential computation by employing a transformer as a replacement for the RNN. RSSM accesses the past indirectly via compression of \(h_{t-1}  \). However, transformer directly accesses the sequence of stochastic states and actions of the past at every time step as \(h_t = f_{transformer} (z_{1:t-1}, a_{1:t-1}) \). Using a modified Dreamer using \(q(z_t|x_t) \) instead of \(q(z_t|x_t, h_t) \) would perform similarly to the original Dreamer as shown by experiment in paper  \cite{chen2022transdreamer}. We use the modified equation for the representation model in our work. Stochastic state model, Image Predictor, Reward Predictor and Discount Predictor all remain the same as Dreamer-v3 for this project.
To summarize:
\begin{itemize}
    \item Deterministic State Model: Replace the current gated recurrent unit with a transformer encoder that is independent of prior deterministic states $h_t$ and dependent of all representation states $z_t$ and actions $a_t$
    \item Representation Model: Adjust belief to omit the deterministic state $h_t$
    \item Trajectory Roll-out: Cap imagined trajectories to 3 per training sample and repay buffer prioritizes higher reward trajectories
    \item Policy Learning: Freeze transformer parameters during training
\end{itemize}

\subsection{Implementation in DreamerV3}
In the implementation of DreamerV3, the team integrated a transformer into the world model architecture, replacing the GRU within the RSSM module. This task was undertaken using JAX and Ninjax, two relatively new programming languages in the field of machine learning. Ninjax is a neural network library built on top of JAX, providing a more streamlined and flexible approach to building complex neural network architectures.

The decision to implement a naive transformer, where the input and output dimensions mirror those of the original GRU, aimed to focus attention on the prior deterministic state and action only. This approach was intended as a baseline to ensure compatibility with DreamerV3's broader codebase. However, the team encountered challenges due to the nascent state of documentation and community knowledge surrounding Ninjax. This led to a slower-than-anticipated development and integration process for the transformer.

In our naive transformer implementation, the attention mechanism is manually constructed with separate linear transformations for queries, keys, and values, followed by a scaled dot-product attention calculation without an attention mask. Our approach omits positional encoding due to only considering a the prior state. In the feed-forward network within each transformer layer, we omit a dropout layer due to conflicts with Ninjax. Layer normalization in our model is applied post-attention and post-feed-forward network addition steps.

\section{Experiments \& Results} \label{experiments_results}

\begin{figure}
\begin{minipage}[c]{0.3\linewidth}
\includegraphics[width=\linewidth]{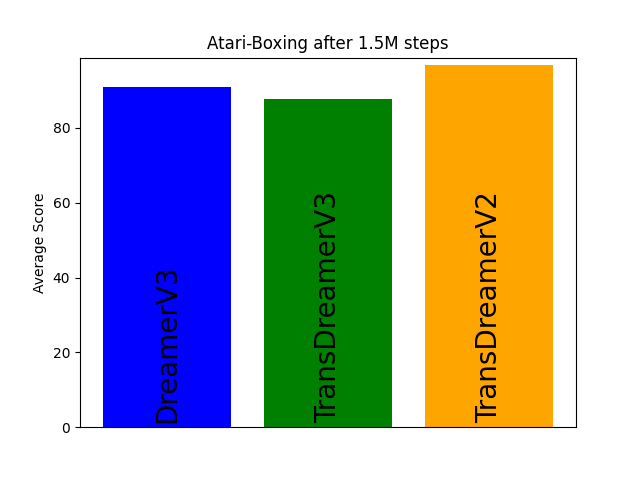}
\end{minipage}
\begin{minipage}[c]{0.3\linewidth}
\includegraphics[width=\linewidth]{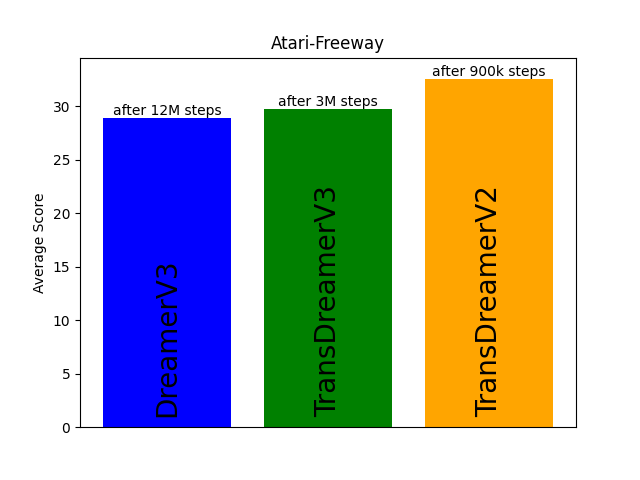}
\end{minipage}
\begin{minipage}[c]{0.3\linewidth}
\includegraphics[width=\linewidth]{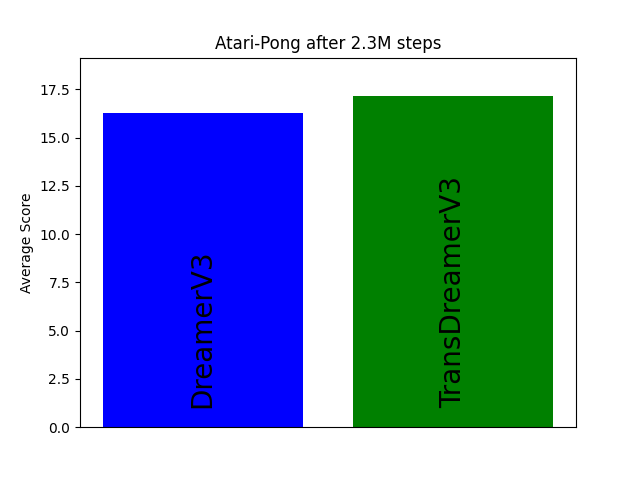}
\end{minipage}
\begin{minipage}[c]{0.3\linewidth}
\includegraphics[width=\linewidth]{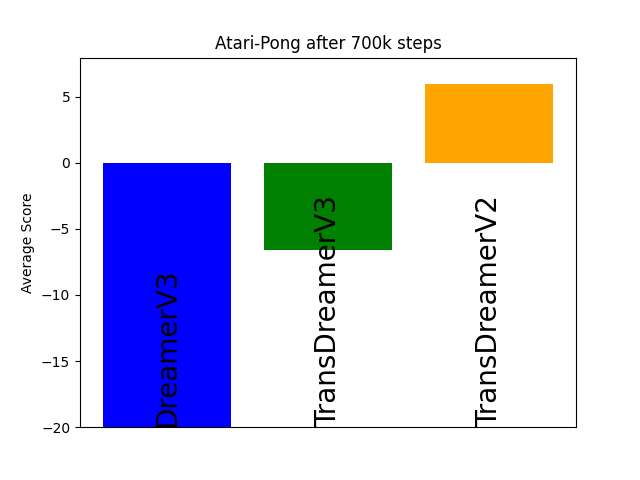}
\end{minipage}
\begin{minipage}[c]{0.3\linewidth}
\includegraphics[width=\linewidth]{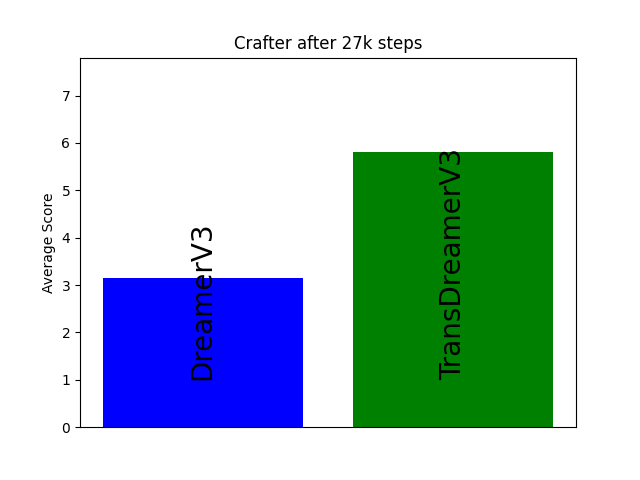}
\end{minipage}%
\caption{Performance comparison of DreamerV3, TransDreamer, and TransDreamerV3}
\label{fig:atari}
\end{figure}

To assess the performance of DreamerV3 \cite{hafner2023mastering}, TransDreamer \cite{chen2022transdreamer}, and our model, we conducted training experiments on Atari-boxing, Atari-freeway, Atari-pong, and Crafter tasks. Due to time and resource constraints, we compare performances at different training steps. Nevertheless, it's important to note that within each task, the comparison is fair as all models are trained under the same settings. Performance comparisons for Atari tasks and Crafter are visualized in Figure~\ref{fig:atari}. 

\subsection{Atari and Crafter Tasks}
For the Atari-Boxing task, TransDreamer \cite{chen2022transdreamer} exhibited superior performance after 1.5 million experimental steps. In the Atari-Freeway task, DreamerV3 \cite{hafner2023mastering} achieved non-zero rewards after 12 million steps, while our model, TransDreamerV3, accomplished the same in 3 million steps. Conversely, TransDreamer \cite{chen2022transdreamer} achieved positive rewards within 1 million steps, surpassing the other two models. In the Atari-Pong task, both DreamerV3 \cite{hafner2023mastering} and our model initially received negative rewards for the first 700k steps. However, our model eventually outperformed DreamerV3 \cite{hafner2023mastering} after 2.3 million steps. Although we didn't complete training for TransDreamer \cite{chen2022transdreamer} up to 2.3 million steps, we will provide updates when available. With the Crafter task, our model significantly outperforms DreamerV3\cite{hafner2023mastering}. In every tested environment except boxing, our model, employing a naive transformer, surpasses DreamerV3 in performance. This outcome indicates that even though both our model and the GRU-based state space model concentrate on the most recent data, the attention mechanism in our model has advantages over the GRU's gated mechanism. However, our model underperforms compared to the original TransDreamer, which aligns with expectations given that our model does not fully utilize the context of previous states.

\subsection{Other experiments}
We encountered several challenges while attempting to run experiments with Minecraft. Despite successfully configuring the settings that yielded results for all other tasks, we consistently faced a runtime error indicating a 'Lost connection to workers.' After resolving setup issues and successfully running the training in Docker, we encountered additional difficulties with the evaluation code. Unfortunately, the issues with the evaluation code prevented us from directly comparing our model with DreamerV3 \cite{hafner2023mastering}. We documented the docker setup for Minecraft since it needed engineering and modifications in the originally provided Dockerfile by \cite{hafner2023mastering}. It's noteworthy that we were able to execute all other experiments presented in DreamerV3 \cite{hafner2023mastering}, including Proprio Control, BSuite, and Visual Control. However, due to resource constraints, we couldn't afford to train models for each task. We have provided code and model checkpoint in the repo$\footnote{https://github.com/XiaonaZhou/TransDreamerV3}$.

\section{Conclusion}\label{conclusion}
Our research introduces TransDreamerV3, a novel reinforcement learning model that integrates the transformer based architecture of TransDreamer into the robust framework of DreamerV3. By replacing the GRU in the RSSM module with a transformer encoder and modifying the belief representation in the model, we aimed to enhance memory capabilities and overall performance. Our model outperformed DreamerV3 in the Atari-Freeway and Crafter tasks, and showed promising results in Atari-Pong suggesting advancement in handling complex environments and decision-making scenarios. Future ablation studies should investigate the GRU gated mechanism compared with the attention mechanism of the transformer. The absence of positional embedding, dropout layers and attention to all representational and action states in our transformer architecture suggests areas for future refinement. 
\section{Contribution}\label{contribution}
We have completed 100\% of this project without any outside help or guidance. Our contribution to this project is as follows: 
\begin{itemize}
  \item Shruti assisted in setting up DreamerV3 on Docker and planning the evaluation \& visualization steps. Worked on running Minecraft experiment. Collaborated to write the related work and a part of Minecraft experiment evaluation.
  \item Amun assisted in implementing the naive transformer in DreamerV3's code base, replacing the GRU. Collaborated to write the introduction, methods and conclusion.
  \item Jonathan led developing the model architecture, planning the overall implementation, and programming \& debugging the naive transformer in DreamerV3's framework. Collaborated to write the introduction, related work, methods and conclusion.
  \item Xiaona took charge of setting up \& training DreamerV3, TransDreamer, TransDreamerV3 for various tasks, getting experimental results, making plots, and providing assistance in implementing the naive transformer. Wrote the experiments and results section, and a part of the related work. 
\end{itemize}

\newpage
{\small

}
\clearpage % Force a new page

\end{document}